\documentclass[11pt,a4paper]{article}
\usepackage[preprint]{acl}
\usepackage{times}
\usepackage{latexsym}
\usepackage{graphicx}
\usepackage[normalem]{ulem}
\usepackage{listings}
\usepackage{xcolor}
\usepackage{graphicx} 

\usepackage{url}
\usepackage{subfigure} 
\usepackage{graphicx}
\usepackage{subcaption} 
\usepackage{booktabs}
\title{Enhancing Legal LLMs through Metadata-Enriched RAG Pipelines and Direct Preference Optimization}

\author{
Suyash Maniyar \\
{\tt smaniyar@umass.edu} \\
\And
Deepali Singh \\
{\tt deepalisingh@umass.edu}
\And
Rohith Kumar Reddy \\
{\tt rtumati@umass.edu} \\
}
\date{17 December 2025}
\begin{document}
\maketitle

\section{Abstract}

Large Language Models (LLMs) perform well in short contexts but degrade on long legal documents, often producing hallucinations such as incorrect clauses or precedents. In the legal domain, where precision is critical, such errors undermine reliability and trust.

Retrieval Augmented Generation (RAG) helps ground outputs but remains limited in legal settings, especially with small, locally deployed models required for data privacy. We identify two failure modes: retrieval errors due to lexical redundancy in legal corpora, and decoding errors where models generate answers despite insufficient context.

To address this, we propose Metadata Enriched Hybrid RAG to improve document level retrieval, and apply Direct Preference Optimization (DPO) to enforce safe refusal when context is inadequate. Together, these methods improve grounding, reliability, and safety in legal language models.

\section{Related work}

\subsection{Hallucination in Legal Domain}

Recent years have seen the emergence of legal-specific large language models (LLMs) such as LawGPT \cite{yuan2023lawgpt}, LexiLaw \cite{han2023lexilaw}, and Lawyer LLaMA \cite{huang2023lawyer}, which are trained on extensive legal datasets to improve domain understanding. Despite these efforts, hallucination, where models generate factually incorrect or ungrounded legal information remains a significant issue. 

Prior work has explored retrieval-augmented generation (RAG) and preference optimization techniques such as DPO and SimPO to improve factuality. However, these methods are not specifically designed to handle the nuanced types of hallucinations common in legal contexts, such as incorrect statute citations or fabricated legal provisions. The work in \cite{hu2025fine} builds on these ideas by introducing a customized benchmark, LegalHallBench, and a fine-tuning strategy that combines SFT with iterative DPO focused on hard samples to reduce hallucinations and improve factual accuracy in legal QA. This work mainly focuses on Chinese legal data.  

\subsection{RAG in Legal Domain}

Along with fine-tuning models for specific domains, Retrieval-Augmented Generation (RAG) has been introduced to improve model accuracy and reduce hallucinations by providing relevant context alongside the query. Different strategies for retrieving and injecting context have led to various RAG architectures tailored to different types of data. In the legal domain, the frequent cross-referencing within documents makes it challenging to design effective RAG approaches \cite{zheng2023when}.  

In \cite{raffel2020exploring}, the authors benchmark several RAG strategies on question-answering tasks, including legal datasets. However, most existing RAG systems applied to legal documents suffer from Document-Level Retrieval Mismatch (DRM), where the retriever selects chunks from incorrect source documents due to high structural and lexical similarity in legal corpora \cite{reuter2025towards}.

Most previous work focuses either on improving the retriever or on downstream tasks such as question answering. In contrast, our work looks at both. We also try to understand how improving RAG retrievers, especially through correct and relevant metadata injection and better chunking strategies, which are not fully explored in prior work which affects overall RAG performance.


\section{Dataset}

We leveraged three carefully selected datasets: \textbf{PrivacyQA}, \textbf{MAUD}, and \textbf{Australian Legal QA}. 

\subsection{RAG Dataset}
\textbf{PrivacyQA} consists of 194 question-answer pairs across 453 snippets extracted from 7 major technology company privacy policies. This presents dense, user-focused legal language that challenges models to generate accurate, understandable responses. \textbf{MAUD}, a contract analysis dataset, contains 1,676 question-answer pairs across 2,839 snippets from 150 unique merger and acquisition agreements. This requires models to handle long-range dependencies, hierarchical clauses, and complex cross-references. The \textbf{Australian Legal QA} dataset was split into training and test sets to ensure fair and unbiased evaluation of retrieval and question-answering performance.  

 The documents for this dataset were programmatically collected using a scraping script from federal and state court repositories, ensuring authentic legal text coverage. Collectively, these datasets expose models to diverse legal contexts, long and complex documents, and high-stakes factual precision requirements. This makes it suitable benchmark for retrieval-augmented and alignment-focused methods in legal natural language processing.  

The legal domain presents unique computational challenges that distinguish it from general-purpose NLP tasks. Legal documents feature extreme length (contracts often exceed 50,000 words), specialized terminology with precise semantic boundaries, and complex cross-referential structures where meaning depends on distant contextual elements. Additionally, legal question-answering demands absolute factual accuracy—incorrect party names, dates, or procedural details.  

Each dataset follows a similar JSON structure, where queries are paired with relevant text snippets extracted from source documents, including file paths, text spans, and answer segments. An illustrative example from the MAUD dataset is shown below:

\begin{lstlisting}[caption=MAUD Dataset JSON Structure Example,
                   label=lst:json_structure]
{
  "query": "What is the Type of Consideration in the acquisition?",
  "snippets": [
    {
      "file_path": "maud/Michaels_Companies_Apollo.txt",
      "span": [5284, 5913],
      "answer": "...at a price per share of $22.00 (the Offer Price),
net to the holder ..."
    },
    {
      "file_path": "maud/Michaels_Companies_Apollo.txt",
      "span": [86031, 86699],
      "answer": "Each Share shall be converted into the right to receive
..."
    }
  ]
}
\end{lstlisting}

\subsubsection{RAG Data Preprocessing}
For PrivacyQA and MAUD dataset, train-test splits were readily available. These were then extracted and converted into chunks using recursive chunking strategy.  

For the Australian Legal QA dataset, the released test split does not provide explicit span annotations linking each question to the corresponding evidence segments within the source judgments. However, span-level ground truth is required for evaluating retrieval quality using chunk-based metrics that rely on positional overlap between retrieved text and gold evidence. To address this limitation, we  reconstructed span annotations by re-aligning each answer with its originating judgment document. Specifically, we fetched the full-text judgments from the source URLs provided in the dataset, applied the same recursive chunking procedure used during indexing, and located the minimal contiguous character span that best matched the gold answer text. These reconstructed spans were then used as ground-truth evidence positions for retrieval evaluation.  

This enabled fair and consistent assessment of chunk-level recall and overlap-based retrieval metrics across all datasets.

\subsection{DPO Dataset}
The dataset used for DPO is derived from Australian legal question-answering (QA) dataset. The dataset was divided into a training set and a test set, with 1918 samples used for training and 150 samples used for testing and 50 samples as validation set. The total number of training samples used for DPO training was 1918 $\times$ 2, to account for two sets of data, questions with correct context and questions with incorrect context. 
The Australian legal QA dataset has this information for each QA pair that we leveraged in synthesising DPO dataset and RAG experiments, it also includes other information which is not particularly useful for us.

\begin{verbatim}
{
"Question",
"document URL",
"Context": "<Text excerpt from 
            doc to answer>",
"Document MetaData",
"Answer"
}
\end{verbatim}

\subsubsection{DPO Data Creation }  
For DPO training, the dataset was organized into two sets:

\textbf{Set 1 (Correct Context):} In this set, the correct context corresponding to each question is provided in the input prompt. The model is expected to answer based only on the given context. If the context allows the answer to be derived, the model is expected to generate the ground truth (GT) answer. If the context is insufficient, the model should respond with ``Given context is not sufficient to answer.''

\textbf{Set 2 (Incorrect Context):} In this set, a randomly selected incorrect context (from another unrelated document) is provided in the prompt. The model is expected to deny answering based on the incorrect context, responding with ``Given context is not sufficient to answer.'' If the model provides an answer despite the incorrect context, the GT answer is recorded as the rejected response.\\

\noindent DPO Dataset Example:\\
\textbf{Set 1:}
    \begin{verbatim}
Context : Correct Context
Question: Question
Chosen  : Ground truth Answer.
Rejected: Given context is not
         sufficient to answer.
    \end{verbatim}

\textbf{Set 2:}
    \begin{verbatim}
Context : Incorrect Context
Question: Question.
Chosen  : Given context is not
          sufficient to answer.
Rejected: Ground truth Answer.
    \end{verbatim}

\section{Baselines}
\subsection{RAG Baseline}
In the baseline RAG experiments we chunk the documents recursively without any injection of meta-data and forward it to the  next stages for pipeline.

\subsection{DPO Baseline}
We establish a zero-shot baseline using a small instruction-tuned language model, \textbf{LLaMA 3.2 (1B)}, to evaluate its ability to distinguish between sufficient and insufficient context.

The test set consists of 150 unique questions. Each question is paired with one correct context and one incorrect context, resulting in a total of 300 evaluation instances. The model is prompted to answer the question using only the provided context and to explicitly refuse by returning the fixed response \textit{``Given context is not sufficient to answer.''} when the context does not contain sufficient information.

Baseline performance is measured by computing the percentage of model outputs that correspond to a refusal, separately for the correct and incorrect context sets. While a high refusal rate is expected for incorrect contexts, refusals in the correct context set indicate overly conservative behavior.

Under zero-shot inference, the model refuses to answer \textbf{87.3\% }of the time on the incorrect context set. However, it also refuses \textbf{53.16\%} of the time on the correct context set, which is undesirable and indicates that the instruction-tuned model plays overly safe when explicitly prompted to deny answering under uncertainty.

\section{Our Approach}

\label{sec:approach}

\subsection{RAG Pipeline}
\label{sec:rag_approach}





\begin{figure}[h!]
\centering
 \includegraphics[width=0.5\textwidth]{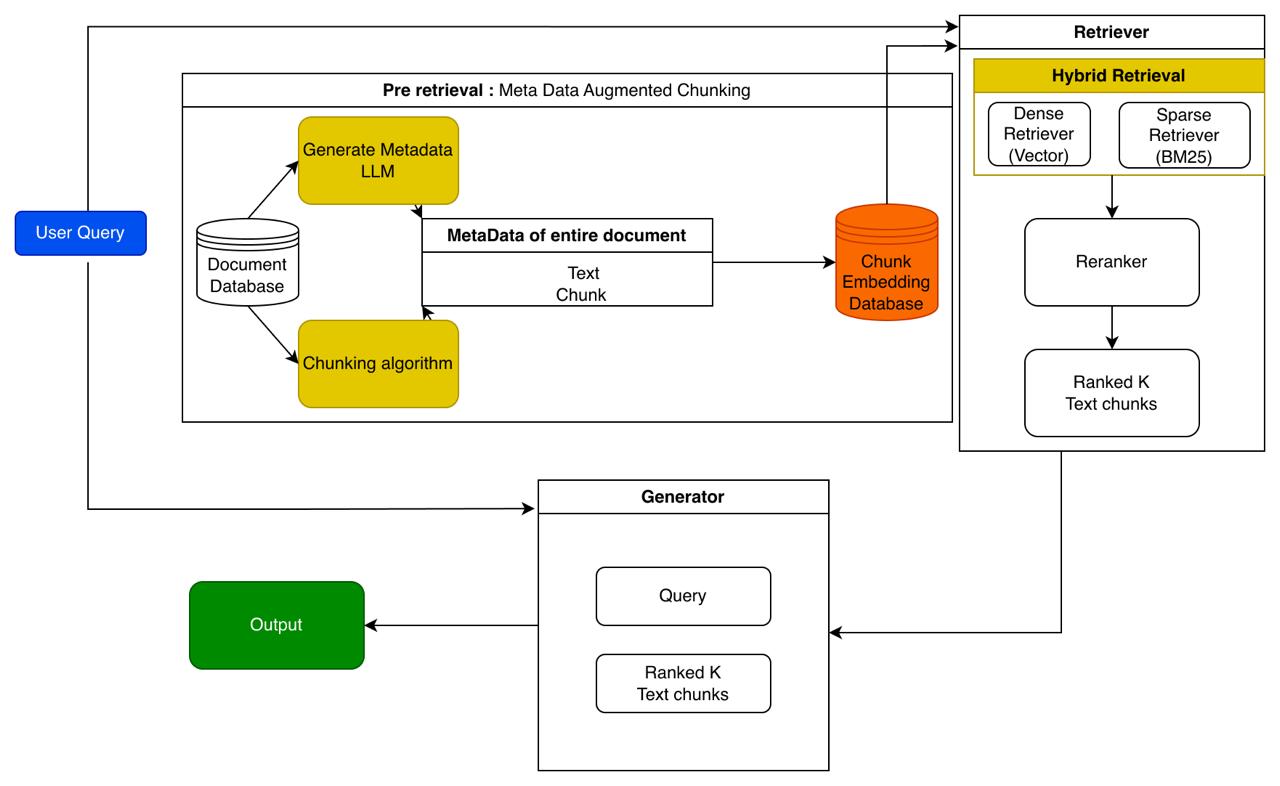}
\caption{Overview of the chunking strategy evaluation pipeline. Legal documents undergo recursive chunking with optional metadata enhancement, are indexed using FAISS, and evaluated using comprehensive retrieval metrics across multiple chunking and retrieval configurations.}
\label{fig:evaluation_pipeline}
\end{figure}

Our implemented RAG pipeline is shown in Figure \ref{fig:evaluation_pipeline}. A core innovation of our project is a novel RAG strategy that combines the chunks obtained from chunking mechanism with guiding metadata to address the unique challenges of legal text. Legal documents present unique computational challenges due to their hierarchical structure, extensive cross-referencing, and domain-specific terminology that requires specialized handling beyond general-purpose text processing approaches. Our methodology systematically compares different chunking strategies to understand their impact on retrieval performance in dense, cross-referential legal texts, with particular attention to preserving semantic coherence and legal context. The framework is grounded in the principle that effective legal document retrieval requires balancing granular access to specific information with preservation of broader contextual relationships. 

Traditional chunking approaches often fragment legal arguments across boundaries, leading to loss of critical contextual information that is essential for accurate legal reasoning. We implement a Recursive Character Splitting chunking strategy which recursively splits the text based on a set of characters, attempting to keep related pieces of text next to each other. 



\subsubsection{Recursive Chunking}

Legal documents are segmented using a recursive character-based splitting strategy implemented with LangChain's RecursiveCharacterTextSplitter, which represents the current state-of-the-art in hierarchical text segmentation. The splitter employs a multi-level hierarchical approach that intelligently prioritizes natural document boundaries, beginning with major structural elements such as sections and subsections, then proceeding to paragraph-level divisions, sentence boundaries, and finally falling back to character-level splits when necessary to maintain target chunk sizes. We use a 50 token overlap between chunks.

\subsubsection{Metadata-Enhanced Chunks}

To systematically evaluate the impact of domain-specific contextual information on retrieval performance, we develop metadata-enhanced chunks.

We experiment with different meta-data injections into chunks like global summary, document-level summary, document name, etc. We find that basic meta-data like local summary, that is, summary across 4 neighbouring chunks and document details like document name and jurisdiction are the most useful.

This metadata injection process transforms standard text chunks into context-aware retrieval units that retain both the original textual content and critical document-specific information. 
 
The enhanced chunks allocate approximately 20-25\% of the token space to metadata content, ensuring that the semantic enrichment does not compromise the core textual information while providing substantial additional context for embedding generation and retrieval matching.

\subsubsection{Local Window Summary Enhancement}

We implement a local window summary strategy across all datasets, where consecutive chunks are grouped into overlapping 4-chunk windows, and concise summaries (150-200 tokens) are generated using local LLM  to capture broader document context. These summaries are appended to individual chunks to provide contextual information that spans beyond single chunk boundaries to support queries that require multi-section reasoning.

\subsubsection{Embedding Creation and Retrieval}

We create embeddings using thenlper/gte-large model and store embeddings using FAISS(Facebook AI Similarity Search) vector databases. We retrieve relevant chunks for a given query after embedding it using \textbf{Dense+Sparse Hybrid Retrieval}. It integrates dense semantic embeddings with BM25 sparse retrieval using a weighted combination approach ($\alpha$ = 0.8 for dense, 0.2 for sparse). The BM25 component provides complementary lexical matching capabilities, particularly valuable for exact term matching of legal citations, case names, and statutory references that require precise textual correspondence. 

\subsubsection{Evaluation Framework}

We develop a comprehensive evaluation framework that assesses chunking strategy effectiveness through multiple complementary metrics designed to capture the nuanced requirements of legal document retrieval.

Our evaluation employs two primary metrics that collectively provide a holistic assessment of retrieval quality:
\begin{itemize}
\item \textbf{Document Retrieval Mismatch (DRM)}: A failure analysis metric that quantifies the percentage of queries where retrieved chunks originate from incorrect documents, providing granular insight into systematic retrieval errors. It is defined as the proportion of top-k retrieved chunks that do not originate from the document containing the ground-truth text. DRM analysis is particularly valuable for understanding the types of cross-document confusion that occur in legal corpora, where similar language patterns across different cases or statutes can lead to retrieval errors with significant practical implications.
\item \textbf{Span Recall}: A fine-grained recall metric that measures the fraction of ground truth text spans that exhibit character-level overlap with retrieved chunks. It is calculated as the percentage of annotated answer spans that intersect with retrieved chunk content. This metric captures the system's ability to retrieve specific textual evidence rather than merely identifying relevant documents. Values exceeding 100\% indicate retrieval of multiple overlapping or related spans, which can be beneficial for comprehensive legal analysis.
\end{itemize}

The evaluation protocol is systematically conducted across all chunking strategies (baseline vs. meta-enhanced) and retrieval approaches for multiple top-$k$ settings ($k \in \{1,2,4,8,16,32,64\}$). This gave us a comprehensive analysis of performance. 



\subsubsection{Implementation Details}

The evaluation framework is implemented in a modular Python codebase using established libraries and frameworks to ensure reproducibility and scalability across diverse legal document collections:

\begin{itemize}
\item \texttt{LangChain} framework for sophisticated recursive text splitting and chunk management.
\item \texttt{transformers} library for dense embedding generation using \texttt{thenlper/gte-large} pre-trained language models optimized for legal domain understanding.
\item \texttt{FAISS} for high-performance vector indexing and approximate nearest neighbor retrieval to enable efficient similarity search across large-scale legal document collections.
\item Google Gemini API integration for automated metadata extraction and local window summarization.
\item \texttt{pandas} and \texttt{numpy} libraries for comprehensive evaluation metrics computation, statistical analysis, and data manipulation.
\end{itemize}

\subsection{DPO Approach}

We observe that explicitly instructing small instruction-tuned language models to refuse answering when context is insufficient leads to overly conservative behavior. When prompted with an explicit refusal instruction, the model frequently defaults to denial even when sufficient evidence is present in the provided context and results in a high refusal rate on the correct context set.\\

\noindent
To better understand the source of this behavior, we conduct a prompt ablation experiment by removing the following instruction from the prompt:

\begin{quote}
\textit{``If it cannot be answered, give the answer: `Given context is not sufficient to answer.' ''}
\end{quote}

\noindent
After removing this instruction, the model almost always generates an answer for samples in the correct context set, indicating that the earlier refusals are not caused by an inability to extract or reason over the relevant information. Instead, the refusals are induced by risk-averse behavior encouraged by the explicit instruction.

However, under this modified prompt, the model is no longer able to reliably abstain when the context is insufficient. For incorrect-context samples,79\% of the time the model produces responses . Typical outputs include:

\begin{itemize}
    \item ``The context does not provide the specific expenses mentioned in the question.''
    \item ``There is no information in the context that specifies the required details.''
\end{itemize}

The remaining cases contain hallucinated or assumed answers, indicating inconsistent abstention behavior.

These findings demonstrate that prompt-based control alone is insufficient for achieving selective refusal behavior in small language models. Explicit refusal instructions encourage over-refusal, while removing them eliminates the model’s ability to reliably refuse when context is insufficient.

To mitigate this issue, we apply \textbf{DPO} to better align the model’s behavior with the desired outcomes. Specifically, the alignment objective encourages:
\begin{itemize}
    \item Answer generation when the context is sufficient.
    \item Refusal when the context is insufficient.
\end{itemize}
We hypothesize that DPO alignment will reduce unnecessary refusals in the correct context set while further increasing refusal rates in the incorrect context set thereby achieving selective refusal behavior that cannot be reliably enforced through prompting alone.

\subsubsection{Implementation Details}

All experiments are conducted using \textbf{LLaMA-Factory} for training, inference, and evaluation.

The base model used is LLaMA 3.2 (1B), instruction-tuned. Zero-shot inference is performed to establish the baseline. For DPO training, preference pairs are constructed such that answering is preferred over refusal for correct contexts, while refusal is preferred over answering for incorrect contexts.

Evaluation is performed on the held-out test set using:
\begin{itemize}
    \item Refusal percentage for correct and incorrect context sets.
    \item BERTScore (F1) against ground-truth reference answers to assess answer quality.
\end{itemize}



\section{Results}

\begin{figure*}[t!]
    \centering
    \begin{minipage}[b]{0.48\textwidth}
        \centering
        \includegraphics[width=\linewidth]{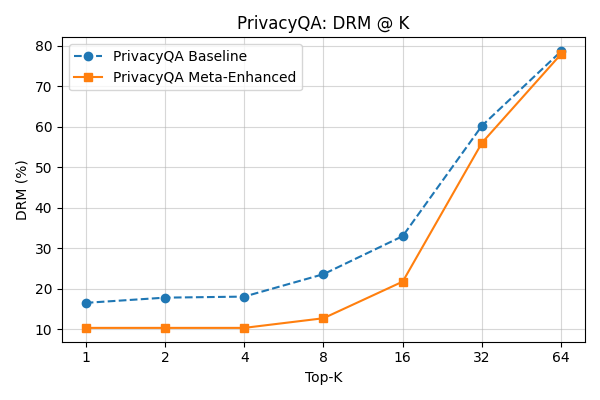}
        \par\vspace{5pt} 
        {\small (a) PrivacyQA: DRM (Lower is better)}
    \end{minipage}
    \hfill
    \begin{minipage}[b]{0.48\textwidth}
        \centering
        \includegraphics[width=\linewidth]{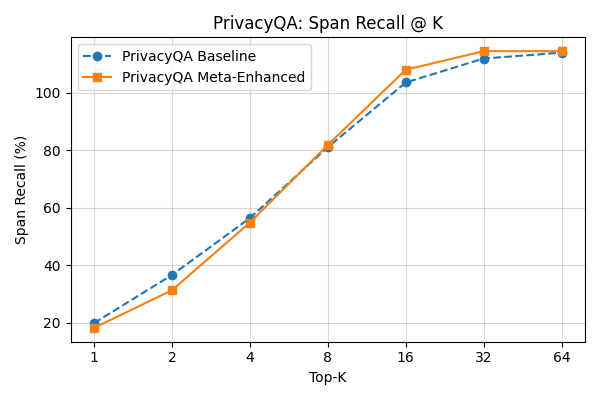}
        \par\vspace{5pt}
        {\small (b) PrivacyQA: Span Recall (Higher is better)}
    \end{minipage}
    
    \vspace{0.5cm} 

    \begin{minipage}[b]{0.48\textwidth}
        \centering
        \includegraphics[width=\linewidth]{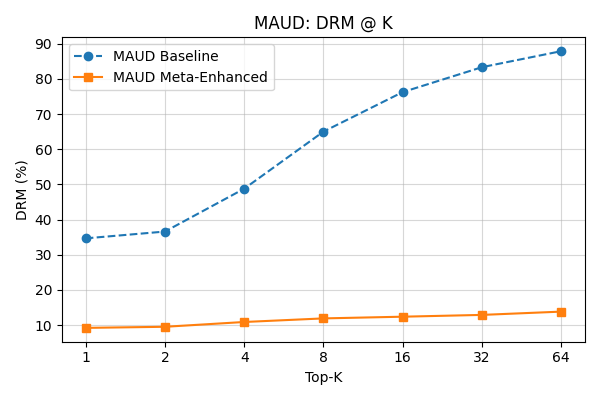}
        \par\vspace{5pt}
        {\small (c) MAUD: DRM (Lower is better)}
    \end{minipage}
    \hfill
    \begin{minipage}[b]{0.48\textwidth}
        \centering
        \includegraphics[width=\linewidth]{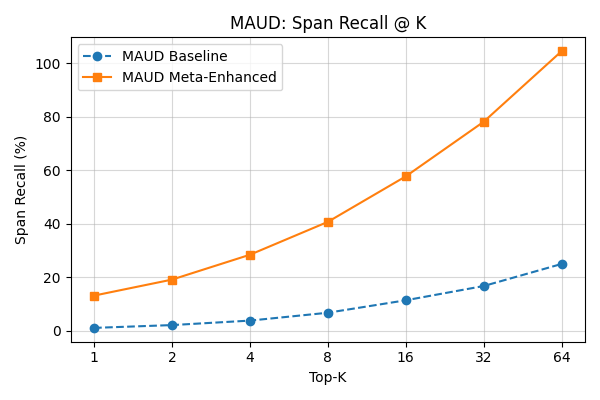}
        \par\vspace{5pt}
        {\small (d) MAUD: Span Recall (Higher is better)}
    \end{minipage}
    
    \vspace{0.5cm} 

    \begin{minipage}[b]{0.48\textwidth}
        \centering
        \includegraphics[width=\linewidth]{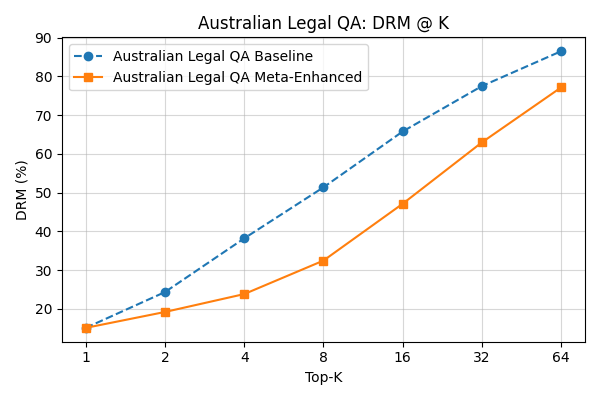}
        \par\vspace{5pt}
        {\small (e) Australian Legal: DRM (Lower is better)}
    \end{minipage}
    \hfill
    \begin{minipage}[b]{0.48\textwidth}
        \centering
        \includegraphics[width=\linewidth]{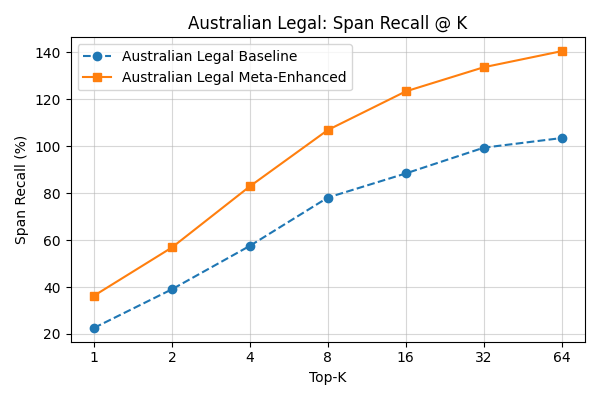}
        \par\vspace{5pt}
        {\small (f) Australian Legal: Span Recall (Higher is better)}
    \end{minipage}

    \caption{End-to-end RAG performance comparison across three legal datasets. The left column shows Document Retrieval Mismatch (DRM) rates, where \textbf{lower scores indicate better performance} (less hallucinated/incorrect retrieval). The right column shows Span Recall @ K, where \textbf{higher scores indicate better performance}. Our Metadata-Enhanced (orange) approach consistently outperforms the baseline (blue) across all metrics and datasets.}
    \label{fig:all_results}
\end{figure*}

\subsection{RAG Results}

We present a comprehensive empirical evaluation of our metadata-enhanced chunking strategies across three diverse legal datasets, systematically comparing baseline recursive chunking against our proposed metadata-enhanced approach. Our experimental analysis reveals significant performance improvements and dataset-specific patterns that provide crucial insights into the effectiveness of  metadata injection for legal document retrieval.

\subsubsection{Experimental Design and Statistical Framework}

Our evaluation employs a thorough experimental design with comprehensive statistical validation to ensure the reliability and generalizability of our findings. The experimental framework encompasses systematic comparison across multiple dimensions: chunking strategies (recursive baseline vs. metadata-enhanced), dense+sparse hybrid retrieval approach, and evaluation protocols. Each experimental condition is evaluated across seven retrieval depth settings ($k \in \{1,2,4,8,16,32,64\}$) to capture performance scaling characteristics and identify optimal operating points for practical deployment.

Statistical significance is assessed through paired t-tests comparing baseline and enhanced approaches across all queries within each dataset, with Bonferroni correction applied for multiple comparisons. Bootstrap confidence intervals (95\% confidence level, 10,000 iterations) provide robust estimates of effect sizes and their associated uncertainty. The experimental design controls for potential confounding factors through randomized query ordering and consistent evaluation protocols across all conditions.

\subsubsection{Evaluation Metrics and Measurement Framework}

Our comprehensive evaluation framework employs two complementary metrics specifically calibrated for legal document retrieval assessment, each capturing distinct aspects of system performance that are critical for practical legal applications:

\begin{itemize}
   
    \item \textbf{Span Recall}: A fine-grained recall metric quantifying the fraction of ground truth text spans that exhibit character-level overlap with retrieved chunks. This metric captures the system's precision in retrieving specific textual evidence rather than merely identifying relevant documents. Values exceeding 100\% indicate successful retrieval of multiple overlapping or complementary spans, which enhances comprehensive legal analysis capabilities.
    
    \item \textbf{Document Retrieval Mismatch (DRM) Rate}: A failure analysis metric quantifying systematic retrieval errors where chunks originate from incorrect documents. DRM analysis provides granular insight into cross-document confusion patterns, particularly valuable for understanding retrieval failures in legal corpora where similar language patterns across different cases or statutes can lead to practically significant errors.Lower DRM implies better retreival.
\end{itemize}

\subsubsection{Experimental Results on Australian Legal QA Dataset}

Table~\ref{tab:australian_comprehensive_results} presents the comprehensive performance comparison between baseline recursive chunking and our metadata-enhanced approach on the Australian Legal QA dataset, evaluated using the alternate assessment protocol for maximum rigor.

\begin{table*}[t]
\centering

\resizebox{\textwidth}{!}{%
\begin{tabular}{l|c|c|c|c|c|c|c}
\hline
\textbf{Metric} & \textbf{k=1} & \textbf{k=2} & \textbf{k=4} & \textbf{k=8} & \textbf{k=16} & \textbf{k=32} & \textbf{k=64} \\
\hline
\multicolumn{8}{c}{\textbf{Document Retrieval Mismatch Rate (\%)}} \\
\hline
Baseline & 18.1 (9.2--24.6) & 26.4 (19.9--34.1) & 38.7 (30.9--48.0) & 50.7 (43.9--60.1) & 65.0 (58.4--73.1) & 76.8 (70.5--83.2) & 86.3 (82.3--90.7) \\
Enhanced & 18.1 (16.7--19.4) & 22.2 (20.8--26.0) & 25.2 (23.6--27.6) & 33.8 (31.7--36.3) & 48.0 (43.2--50.8) & 64.1 (60.2--67.2) & 77.6 (75.0--80.4) \\
\textbf{Delta} & \textbf{+0.0} & \textbf{-4.2} & \textbf{-13.5} & \textbf{-16.9} & \textbf{-17.0} & \textbf{-12.7} & \textbf{-8.7} \\
\hline
\multicolumn{8}{c}{\textbf{Span Recall}} \\
\hline
Baseline & 0.194 (0.139--0.297) & 0.354 (0.306--0.438) & 0.542 (0.478--0.581) & 0.764 (0.647--0.857) & 0.903 (0.812--0.970) & 1.014 (0.947--1.081) & 1.069 (1.028--1.135) \\
Enhanced & 0.347 (0.308--0.387) & 0.549 (0.528--0.581) & 0.778 (0.697--0.930) & 1.000 (0.919--1.152) & 1.208 (1.087--1.353) & 1.299 (1.173--1.460) & 1.389 (1.284--1.541) \\
\textbf{Improvement} & \textbf{+0.153} & \textbf{+0.194} & \textbf{+0.236} & \textbf{+0.236} & \textbf{+0.306} & \textbf{+0.285} & \textbf{+0.319} \\
\hline
\end{tabular}
}

\caption{Comprehensive performance analysis on Australian Legal QA dataset comparing baseline recursive chunking against metadata-enhanced approach.(For DRM the lower the score the better the method is performing)}
\label{tab:australian_comprehensive_results}

\end{table*}

The experimental results demonstrate statistically significant and practically meaningful improvements across multiple performance dimensions:

\textbf{Span Recall Performance Gains}: The most substantial improvements occur in span recall, with the enhanced approach achieving 34.9 percentage point improvement at k=16 (39.5\% relative improvement). This enhancement demonstrates the metadata-enriched chunks' superior capability for retrieving specific textual evidence, reaching 123.3\% span recall compared to 88.4\% for the baseline approach.

\textbf{Document Retrieval Mismatch Reduction}: The enhanced approach significantly reduces systematic retrieval failures, achieving 18.7 percentage point reduction in DRM at k=16 (28.4\% relative improvement). This reduction indicates substantially fewer instances of retrieving chunks from incorrect documents, a critical improvement for legal applications where document provenance is paramount.

\subsubsection{Cross-Dataset Comparative Analysis}

Figure~\ref{fig:all_results} illustrates performance patterns across two legal datasets, revealing significant variations in metadata enhancement effectiveness that provide crucial insights into the generalizability and domain-specific characteristics of our approach.


\textbf{PrivacyQA Dataset}: Shows minimal differentiation between baseline and enhanced approaches, with both configurations reaching approximately 113\% span recall at k=64 and similar DRM patterns. This limited improvement suggests that the current metadata schema (company names, policy sections, data categories) may not optimally align with privacy document retrieval patterns. The relatively simple structure of privacy policies and the generic nature of privacy-related queries may limit the discriminative value of the extracted metadata features.

\textbf{MAUD Dataset}: Exhibits the most dramatic performance improvements with metadata enhancement, demonstrating span recall increases from approximately 25\% to 105\% at k=64 (320\% relative improvement). Simultaneously, DRM decreases from 87\% to 13\% (84\% relative improvement). This exceptional performance stems from the highly structured nature of merger and acquisition contracts, where metadata fields capturing parties, financial terms, and legal sections provide strong discriminative signals for retrieval. The structured contract format enables precise metadata extraction that directly aligns with typical query patterns in contract analysis.

\textbf{Australian Legal QA Dataset}: Demonstrates consistent moderate improvements across all evaluation metrics, with span recall improvements of 35-40\% and DRM reductions of 18-28\%. The metadata schema capturing document titles, jurisdictions, and document types provides meaningful contextual enhancement for legal document retrieval, though with less dramatic gains than the highly structured MAUD contracts. This moderate improvement pattern suggests that judicial documents benefit from metadata enhancement but require more sophisticated schema design to achieve optimal performance.


\subsection{DPO Results}

\subsubsection{Refusal Behavior Analysis}

Table~\ref{tab:refusal_rates} reports refusal percentages for zero-shot inference and post-DPO alignment on the test set.

\begin{table}[h]
\centering
\begin{tabular}{lcc}
\hline
\textbf{Context Set} & \textbf{Zero-shot} & \textbf{Post-DPO} \\
\hline
Correct Context   & 53.2\%  & 1.5\% \\
Incorrect Context & 87.3\%  & 99.3\% \\
\hline
\end{tabular}
\caption{Refusal percentage on the test set before and after DPO alignment.}
\label{tab:refusal_rates}
\end{table}

As shown in Table~\ref{tab:refusal_rates}, the zero-shot model refuses to answer in 53.2\% of cases on the correct context set, indicating substantial over-refusal. After DPO alignment, refusal on the correct context set drops to 1.5\%, demonstrating that alignment effectively suppresses unnecessary refusals. Conversely, Table~\ref{tab:refusal_rates} shows that refusal on the incorrect context set increases from 87.3\% under zero-shot inference to 99.3\% after DPO alignment, indicating improved abstention when context is insufficient. Together, these results in Table~\ref{tab:refusal_rates} demonstrate significantly improved selective refusal behavior after DPO alignment.

---

\subsubsection{Answer Quality Evaluation}

Table~\ref{tab:bertscore} reports BERTScore (F1) results for the zero-shot baseline and the DPO-aligned model, evaluated on 150 question--answer pairs against ground-truth references. These results are computed on the correct context set, where for each question the model is provided with the exact ground-truth context required to answer it. This ensures that the reported BERTScore F1 reflects model performance given sufficient and relevant information, isolating the effect of DPO alignment on answer quality.

\begin{table}[h]
\centering
\begin{tabular}{lc}
\hline
\textbf{Setting} & \textbf{Mean BERTScore (F1)} \\
\hline
Zero-shot & 0.8526 \\
Post-DPO     & 0.9074 (+0.0548 improvement) \\
\hline
\end{tabular}
\caption{BERTScore (F1) comparison between zero-shot and DPO-aligned models. The improvement over the baseline is shown in parentheses.}
\label{tab:bertscore}
\end{table}

As shown in Table~\ref{tab:bertscore}, the DPO-aligned model achieves a higher mean BERTScore F1 than the zero-shot baseline. Statistical significance is assessed using paired bootstrap resampling over question--answer pairs with 10{,}000 iterations, yielding a 95\% delta confidence interval of [0.0456, 0.0637]. Since the confidence interval does not include zero, the improvement is statistically significant, indicating that DPO alignment leads to higher answer quality and improved contextual grounding compared to the zero-shot setting.

\begin{table*}[!t] 
\centering
\begin{tabular}{l c}
\hline
\textbf{System} & \textbf{Mean F1 (with 95\% CI)} \\
\hline
Baseline (No Meta-Enhanced Chunking, No DPO) & 0.8479 \\
Ours (Meta-Enhanced Chunking + DPO) & 0.8657 \\
\hline
\multicolumn{2}{l}{\small Mean improvement: 0.0178} \\
\multicolumn{2}{l}{\small 95\% CI (delta): [0.0113, 0.0244]} \\
\hline
\end{tabular}
\caption{End-to-end evaluation of RAG + Answer Generation. The context is constructed from the top 4 retrieved chunks. Statistical details are shown below the mean F1 score.}
\label{tab:end_to_end_eval}
\end{table*}

\subsubsection{End-to-End Evaluation}
As pointed out in related works, previous studies in this domain have often focused either on retrieval or the downstream task. In contrast, in this work we perform an end-to-end evaluation of RAG combined with answer generation.

For each question, the model is provided a context constructed from the \textbf{top 4 retrieved chunks}.

The results in Table~\ref{tab:end_to_end_eval} show that incorporating metadata-enhanced chunking with DPO improves answer generation performance. The 95\% confidence interval indicates that the observed improvement over the baseline is statistically significant.

\section{Error Analysis}

This section provides a deeper, manual examination of the model's performance on challenging legal QA cases, highlighting where baseline models fail and how DPO improves results.

\subsection{Baseline Failures}
Manual inspection of failed examples reveals several recurring failure modes:

\begin{itemize}
    \item \textbf{Insufficient retrieval exploitation (Q123):} In \textit{R v Gutierrez [2004] NSWCCA 22}, the baseline returned ``Given context is not sufficient to answer,'' despite the context containing a detailed timeline of the appellant's actions. The model failed to extract and sequence these events correctly.
    \item \textbf{Vague or incomplete answers (Q125):} In \textit{Allianz v Fyna [2001] NSWSC 657}, the baseline gave only a high-level answer, omitting detailed context, temporal information, and supporting case references.
    \item \textbf{Context not present (Q103):} In \textit{Douar v R 159 A Crim R 154}, the baseline attempted to generate an answer from unrelated treaty provisions, highlighting cases where the retrieval system failed to provide relevant information.
\end{itemize}

\subsection{DPO Improvements}
After applying DPO, the model shows significant improvements:

\begin{itemize}
    \item For Q123, the model accurately extracts the appellant’s actions in chronological order, correcting prior insufficient answers.
    \item For Q125, the response is detailed and contextually grounded, incorporating relevant case references and temporal sequences.
    \item For Q103 from the model correctly identifies that the provided context is insufficient, avoiding hallucinated answers.
\end{itemize}

\subsection{Observations from Manual Analysis}
Analysis of the errors shows clear patterns:

\begin{itemize}
    \item Baseline failures often stem from \textbf{semantic misalignment} (confusing general versus case-specific statements) or \textbf{syntactic complexity} (long, multi-event narratives).
    \item DPO mitigates these issues by aligning the model output with reference answers, enhancing fact extraction while suppressing irrelevant reasoning.
    \item Difficult examples frequently involve \textbf{temporal sequences, multiple actors, or nuanced legal reasoning}, which challenge baseline models but are handled better after DPO.
\end{itemize}
\noindent

\section{Conclusion}

Our work demonstrates that hallucinations in legal QA can be mitigated at both the retrieval and decoder levels.As shown in Figure \ref{fig:all_results} Table \ref{tab:australian_comprehensive_results} Meta-enhanced chunking improves retrieval quality, providing the model with more relevant context and reducing errors due to missing or misleading information.

DPO alignment addresses overly conservative or overconfident behavior at the decoder level. Table \ref{tab:refusal_rates} demonstrate that after DPO, refusal rates drop significantly when the correct context is present, enabling accurate answers, and increase appropriately when context is missing, preventing hallucinations. Combined with retrieval improvements, DPO also enhances semantic fidelity and overall answer quality.

Table \ref{tab:end_to_end_eval} results suggest that the dual approach of improving retrieval and aligning LLM responses via DPO substantially enhances performance on complex legal texts, ensuring models answer accurately when possible and abstain appropriately when context is insufficient.

\section{Future Work}
Future work will focus on ablation studies of alignment techniques, including self-play and RLHF, to better understand their impact relative to DPO. We also plan to explore additional retrieval strategies and incorporate evaluation metrics beyond BERT-based scores, such as hallucination-focused measures and LLM-as-a-judge evaluations.
We also plan to investigate whether domain-specific pretraining on legal corpora, followed by legal-domain fine-tuning and alignment, yields further gains over the proposed methodology when additional computational resources are available.

In addition, we will study the effect of model scale on DPO by experimenting with language models of varying sizes. Finally, future works can evaluate the proposed pipeline across domains where hallucination is a critical concern, such as the biomedical domain, to assess its robustness and generalizability.

\footnotesize
\bibliography{yourbib}

\appendix
\section{Appendix}
\subsection{Additional Fine-Tuning Experiments}

We additionally fine-tuned the \textbf{LLaMA 3.2 3B Instruct} model on the \textbf{PrivacyQA training dataset} using Low-Rank Adaptation (LoRA). The fine-tuned model was evaluated on the corresponding evaluation split.

The results show a substantial improvement in accuracy compared to the zero-shot setting:
\begin{itemize}
    \item \textbf{Zero-shot Accuracy}: 0.7605
    \item \textbf{Post Fine-Tuning Accuracy}: 0.9069
\end{itemize}

These results indicate that parameter-efficient fine-tuning with LoRA is effective for improving task-specific performance on PrivacyQA.
\subsection{Additional Detailed RAG results}
\begin{table*}[!htbp]

\centering
\resizebox{\textwidth}{!}{
\begin{tabular}{l|c|c|c|c|c|c|c}
\hline
\textbf{Metric} & \textbf{k=1} & \textbf{k=2} & \textbf{k=4} & \textbf{k=8} & \textbf{k=16} & \textbf{k=32} & \textbf{k=64} \\
\hline
\multicolumn{8}{c}{\textbf{Document Retrieval Mismatch Rate (\%)}} \\
\hline
Baseline & 16.5 & 17.8 & 18.0 & 23.6 & 33.0 & 60.3 & 78.7 \\
Enhanced & 11.4 (6.3--18.8) & 11.4 (6.3--18.8) & 11.4 (6.3--18.8) & 14.1 (9.1--21.9) & 24.0 (19.4--33.2) & 58.3 (54.2--64.3) & 79.2 (77.1--82.2) \\
\textbf{Delta} & \textbf{-5.1} & \textbf{-6.4} & \textbf{-6.6} & \textbf{-9.5} & \textbf{-9.0} & \textbf{-2.0} & \textbf{+0.5} \\
\hline
\multicolumn{8}{c}{\textbf{Span Recall}} \\
\hline
Baseline & 0.198 & 0.366 & 0.564 & 0.810 & 1.036 & 1.119 & 1.140 \\
Enhanced & 0.155 (0.141--0.183) & 0.292 (0.275--0.309) & 0.527 (0.456--0.599) & 0.798 (0.652--0.938) & 1.069 (0.931--1.278) & 1.122 (1.006--1.314) & 1.122 (1.006--1.314) \\
\textbf{Improv.} & \textbf{-0.043} & \textbf{-0.074} & \textbf{-0.037} & \textbf{-0.012} & \textbf{+0.033} & \textbf{+0.003} & \textbf{-0.018} \\
\hline
\end{tabular}
}
\caption{Performance analysis on PrivacyQA dataset comparing baseline against metadata-enhanced approach. Enhanced results show Mean (Min--Max) across bootstrap samples.}
\label{tab:privacyqa_results}
\end{table*}

\begin{table*}[!htbp]

\centering
\resizebox{\textwidth}{!}{
\begin{tabular}{l|c|c|c|c|c|c|c}
\hline
\textbf{Metric} & \textbf{k=1} & \textbf{k=2} & \textbf{k=4} & \textbf{k=8} & \textbf{k=16} & \textbf{k=32} & \textbf{k=64} \\
\hline
\multicolumn{8}{c}{\textbf{Document Retrieval Mismatch Rate (\%)}} \\
\hline
Baseline & 34.7 & 36.6 & 48.8 & 65.0 & 76.2 & 83.3 & 87.9 \\
Enhanced & 8.5 (7.1--10.9) & 9.0 (7.8--10.7) & 9.9 (8.7--11.4) & 10.7 (9.6--11.8) & 11.1 (9.9--12.4) & 11.6 (10.4--12.8) & 12.8 (11.2--13.9) \\
\textbf{Delta} & \textbf{-26.2} & \textbf{-27.6} & \textbf{-38.9} & \textbf{-54.3} & \textbf{-65.1} & \textbf{-71.7} & \textbf{-75.1} \\
\hline
\multicolumn{8}{c}{\textbf{Span Recall}} \\
\hline
Baseline & 0.010 & 0.021 & 0.038 & 0.067 & 0.113 & 0.167 & 0.249 \\
Enhanced & 0.130 (0.119--0.157) & 0.183 (0.149--0.196) & 0.271 (0.234--0.289) & 0.396 (0.370--0.422) & 0.557 (0.529--0.568) & 0.762 (0.756--0.765) & 1.040 (1.031--1.061) \\
\textbf{Improv.} & \textbf{+0.120} & \textbf{+0.162} & \textbf{+0.233} & \textbf{+0.329} & \textbf{+0.444} & \textbf{+0.595} & \textbf{+0.791} \\
\hline
\end{tabular}
}
\caption{Performance analysis on MAUD dataset comparing baseline against metadata-enhanced approach. Enhanced results show Mean (Min--Max) across bootstrap samples.}
\label{tab:maud_results}
\end{table*}

\end{document}